\documentclass[letterpaper, 10 pt, conference]{ieeeconf}  % Comment this line out if you need a4paper

\IEEEoverridecommandlockouts                              % This command is only needed if 
\usepackage{graphics} % for pdf, bitmapped graphics files
\usepackage{epsfig} % for postscript graphics files
\usepackage{times} % assumes new font selection scheme installed
\usepackage{amsmath} % assumes amsmath package installed
\usepackage{amssymb}  % assumes amsmath package installed
\usepackage{booktabs}
\usepackage{bm}
\usepackage{xcolor}
\usepackage{bbm}
\usepackage{hyperref}
\usepackage{pifont}
\usepackage{array}
\usepackage{amsfonts}
\usepackage[normalem]{ulem}

\newcommand{\R}{\mathbb{R}}
\newcommand{\norm}[1]{\left\lVert#1\right\rVert}
\newcommand{\cmark}{\ding{51}}%
\newcommand{\xmark}{\ding{55}}%
\newcolumntype{P}[1]{>{\centering\arraybackslash}p{#1}}

\def\1{\mathbbm{1}}

\title{\LARGE \bf
QuadSwarm: A Modular Multi-Quadrotor Simulator for Deep Reinforcement Learning with Direct Thrust Control
}

\author{Zhehui Huang, Sumeet Batra, Tao Chen, Rahul Krupani, Tushar Kumar \\ 
Artem Molchanov, Aleksei Petrenko, James A. Preiss, Zhaojing Yang, Gaurav S. Sukhatme  % <-this % stops a space
\thanks{The authors did this work at the Department of Computer Science, University of Southern California, Los Angeles, CA 90089 USA (e-mail: zhehuihu@usc.edu). GSS holds concurrent appointments as a Professor at USC and as an Amazon Scholar. This paper describes work performed at USC and is not associated with Amazon.}% <-this % stops a space
}

\begin{document}

\maketitle
\thispagestyle{empty}
\pagestyle{empty}

%%%%%%%%%%%%%%%%%%%%%%%%%%%%%%%%%%%%%%%%%%%%%%%%%%%%%%%%%%%%%%%%%%%%%%%%%%%%%%%%
\begin{abstract}
Reinforcement learning (RL) has shown promise in creating robust policies for robotics tasks.
However, contemporary RL algorithms are data-hungry, often requiring billions of environment transitions to train successful policies. This necessitates the use of fast and highly-parallelizable simulators. In addition to speed, such simulators need to model the physics of the robots and their interaction with the environment to a level acceptable for transferring policies learned in simulation to reality.  
%{\color{red}{Besides, sim-to-real gap degrades the performance of the trained policies when deploy them to the real world. This necessitates accurate physics simulation and well-modeled environment dynamics.}} 
% Unmodeled environment dynamics and simplifying physics assumptions of robotics simulators lead to a sim2real gap. 
We present QuadSwarm, a fast, reliable simulator for research in single and multi-robot RL for quadrotors that addresses both issues. QuadSwarm, with fast forward-dynamics propagation decoupled from rendering, is designed to be highly parallelizable such that throughput scales linearly with additional compute.
It provides multiple components tailored toward multi-robot RL, including diverse training scenarios, and provides domain randomization to facilitate the development and sim2real transfer of multi-quadrotor control policies. Initial experiments suggest that QuadSwarm achieves over 48,500 simulation samples per second (SPS) on a single quadrotor and over 62,000 SPS on eight quadrotors on a 16-core CPU. Code: \href{https://github.com/Zhehui-Huang/quad-swarm-rl}{https://github.com/Zhehui-Huang/quad-swarm-rl}

\end{abstract}

% \james{Overall this paper uses a lot of non-neutral phrases like
% ``proper'',
% ``decent'',
% ``lock the potential''
% }

%%%%%%%%%%%%%%%%%%%%%%%%%%%%%%%%%%%%%%%%%%%%%%%%%%%%%%%%%%%%%%%%%%%%%%%%%%%%%%%%
\section{INTRODUCTION}
Deep reinforcement learning (RL) has shown promise in developing agile control policies for quadrotors~\cite{xie2023learning}. 
However, RL algorithms require a large number of environment transitions to train successful policies in simulation. This motivates building fast and highly-parallelizable simulators. Additionally, it is important for the simulator to be good enough that policies trained on it transfer to the real world in spite of unmodeled environment dynamics and the simplified physics assumptions it will inevitably entail. 
%To deploy policies trained in the simulator to the real world, there is a problem known as sim2real gap. 
%There are a lot of reasons can lead to the sim2real gap problem, such as the unmodeled environment dynamics and the simplified physics assumptions of robotics simulators. 

We describe a simulator, QuadSwarm, to facilitate research in single and multi-robot RL for quadrotors that addresses the aforementioned issues.
Specifically, QuadSwarm supports five main ingredients required to enable the development of RL control policies for real quadrotors: 
$(i)$ A reasonably accurate physics model of a popular existing hardware platform, Crazyflie 2.x, and sufficient domain randomization to account for unmodeled effects;
$(ii)$ Supports per-rotor thrust control;
% with support for per-rotor thrust control and sufficient domain randomization to account for unmodeled effects; 
$(iii)$ Fast single-threaded throughput, highly parallelizable, and scales with additional compute; 
$(iv)$ A diverse collection of learning scenarios for single and multi-quadrotor teams;
$(v)$ 100$\%$ written in Python, which simplifies further development and experimentation.

\begin{figure}[htbp!]
\centering
    % \vspace{-6pt}
    \includegraphics[width=0.46\textwidth]{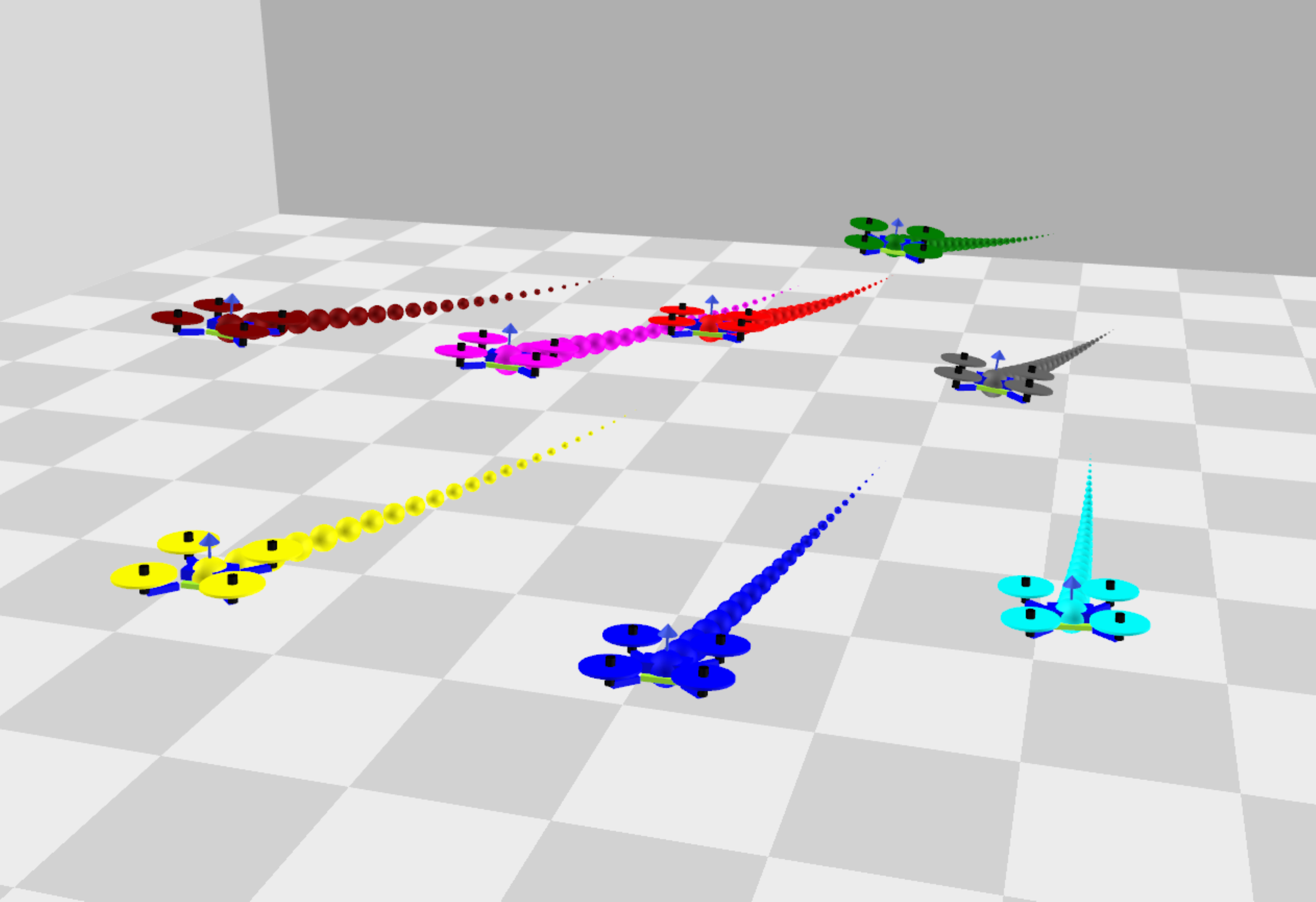}
    \vspace{-6pt}
    \caption{QuadSwarm visualization: 8 quadrotors heading toward a common goal location}
    \label{fig:showcase}
    \vspace{-20pt}
\end{figure}

% To obtain a simulator that meets our requirements, we propose a novel quadrotor simulator: QuadSwarm, which 
% $(i)$ performs fast forward-dynamics propagation based on the Crazyflie physics parameters;
% $(ii)$ is highly parallelizable and scales with additional compute; 
% $(iii)$ supports per-rotor thrust control and domain randomization for facilitating sim2real transfer, and provides a diverse range of difficult training scenarios for single and multi-agent learning.

We evaluate the speed of QuadSwarm on a machine with AMD Ryzen 7 2700X CPU (16 CPU cores). QuadSwarm achieves $>$48,500 simulation samples per second (SPS) in an environment with a single quadrotor and $>$62,000 SPS in an environment with eight quadrotors, enabling collision simulation. In the environment with eight quadrotors, QuadSwarm receives eight samples per simulation step, which speeds up simulation even though additional computation is required for collision. We have demonstrated zero-shot transferability of RL control policies onto real hardware utilizing QuadSwarm in a single~\cite{molchanov2019sim} and multi-quadrotor~\cite{batra2022decentralized} scenarios. 
% In the environment with eight quadrotors, we receive eight simulation samples per simulation step. 
% Therefore, the environment with eight quadrotors achieves higher FPS than single quadrotor even enabling collision simulation.  

% \james{How to compare these? Is the multi-quadrotor FPS something like (num quads) * (dynamics steps per second)? Otherwise it doesn't make sense that it would be higher. Please clarify.}

% One paper focuses on the single quadrotor. It trains one general control policy to stabilize multiple different quadrotors with domain randomization. 
% The other paper focuses on multiple quadrotors. It trains one general control policy to control multiple quadrotors by approaching their own goals and avoiding collisions.

\begin{table*}[t]
  \centering
  \vspace{-12pt}
    \caption{Features Comparing QuadSwarm with Other Simulators that Applicable for Deep RL Research}
    \vspace{-6pt}
    \begin{tabular}{l c c c c c c c}
    \hline
    &\multicolumn{1}{c}{Physics}&&\multicolumn{1}{c}{Supports}&\multicolumn{1}{c}{Per-rotor}&\multicolumn{3}{c}{Multi-agent}\\
    % \hline
    \cline{6-8}
    Simulator & Dynamics & Rendering & Crazyflie & Thrust & Gym Wrapper & Unified Reward Func & Zero-shot Transfer \\
    \hline
    AirSim~\cite{shah2018airsim} & FastPhysicsEngine & UE4 / Unity & \xmark & \xmark & \xmark & \xmark & Wait to Verify \\
    Air Learning~\cite{krishnan2021air} & FastPhysicsEngine & UE4 & \cmark & \xmark & \xmark & \xmark & Wait to Verify \\
    GymFC~\cite{koch2019reinforcement} & Gazebo & OGRE & \xmark & \cmark & \xmark & \xmark & Wait to Verify \\
    Flightmare~\cite{song2021flightmare} & Ad hoc & Unity & \xmark & \cmark & \xmark & \xmark & Wait to Verify \\
    % \cline{1-7}
    gym-pybullet-drones~\cite{panerati2021learning} & PyBullet & OpenGL & \cmark & \cmark & \cmark & \xmark & Wait to Verify\\
    \hline
    \textbf{QuadSwarm} & Ad hoc & OpenGL & \cmark  & \cmark & \cmark & \cmark & \cmark \\
    \hline
    \end{tabular}
    % \vspace{-5pt}

  \label{tab:sim_comparision}
\vspace{-16pt}
\end{table*}

\section{Related Work}
%We discuss existing open-source platforms that support quadrotor simulation. 
%However, they lack some of the aforementioned ingredients necessary for the development of RL-based control policies for real hardware. 

\subsection{Open-source Simulators that Support Single-robot RL}

\subsubsection{AirSim and Air Learning}
AirSim~\cite{shah2018airsim} is a photo-realistic simulator for multiple vehicles, such as cars or quadrotors. However, there are three main limitations of using AirSim in RL research. First, AirSim's physics simulation is coupled with rendering, which limits its simulation speed and parallelization ability. Second, although AirSim supports multiple quadrotors, the physical simulation of collisions is overly simplified~\cite{panerati2021learning}. This makes AirSim unsuitable for control tasks. Third, AirSim does not provide OpenAI Gym~\cite{brockman2016openai} interface for multiple quadrotors. Air Learning~\cite{krishnan2021air}, based on AirSim, focuses on system-level design to address the challenges of training RL policies and deploying them to resource-constrained quadrotors. Air Learning makes AirSim a better fit for learning by addressing several limitations of AirSim, such as using an environment generator to increase the generalization ability of trained policies. However, Air Learning still inherits the three main limitations of AirSim, mentioned above. Different from QuadSwarm, AirSim and Air Learning do not support direct per-rotor thrust control.

\subsubsection{GymFC}
GymFC~\cite{koch2019reinforcement} focuses on tuning flight controllers and developing neuro-flight controllers via RL and supports per-rotor thrust control. While well-suited for developing and tuning single-robot controllers, there is very little support for multi-robot control policies and a lack of a diverse set of training scenarios for multi-robot teams.

\subsubsection{Flightmare}
Flightmare~\cite{song2021flightmare} balances simulation speed, photo-realism, and physical accuracy. It supports a large multi-modal sensor suite and supports two control modes: collective thrust and body rates, and per-rotor thrust. However, Flightmare does not directly support multi-robot RL. 

\subsection{Open-source Simulators that Support Multi-robot RL}
To the best of our knowledge, gym-pybullet-drones~\cite{panerati2021learning} is the only multi-drone simulator besides QuadSwarm that facilitates Deep RL research and development of multi-quadrotor teams. Compared with gym-pybullet-drones, QuadSwarm has three main features that gym-pybullet-drones does not have. First, QuadSwarm implements diverse training scenarios and provides a unified reward function for these scenarios, which increases the generalization ability of trained policies. 
Second, in multi-robot environments, QuadSwarm uses interaction-related rewards, such as the reward when two quadrotors collide with each other. The interaction-related rewards can provide extra information, besides post-collision dynamics, to quadrotors to learn collision avoidance behaviors.
Third, QuadSwarm simulates non-ideal motors and sensor noise to decrease the sim2real gap. 
Besides, in a multi-robot environment with $N$ quadrotors, QuadSwarm uses the relative position and 
 the relative velocity of a fixed number $K$ of nearest robots to represent the neighbor information, where $K\ll N$ when N is large, such as 128, while gym-pybullet-drones uses a boolean distance adjacency matrix $A\in\R^{N \times N}$. Compared with policies trained in gym-pybullet-drones, policies trained in QuadSwarm are thus more easily scalable to larger teams.
 %and able to apply to any number of robots as long as the number is bigger than $K$.}} 

\section{QuadSwarm}

\begin{figure}[htbp!]
\centering
    % \vspace{-2pt}
    \includegraphics[width=0.36\textwidth]{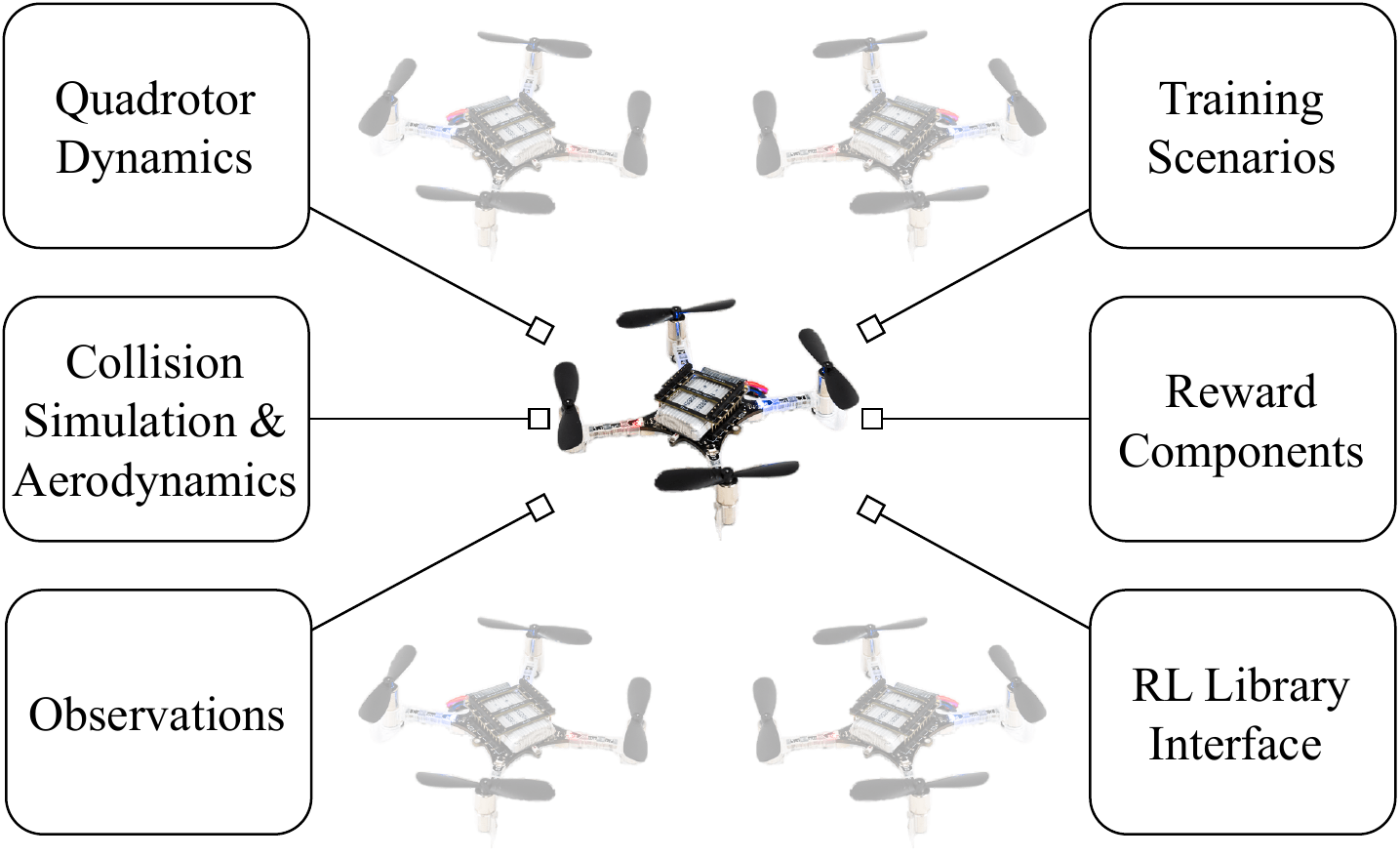}
    \vspace{-8pt}
    \caption{QuadSwarm Simulator Overview 
    }
    \label{fig:system_overview}
    \vspace{-18pt}
\end{figure}

QuadSwarm is a modular quadrotor simulator that supports multiple quadrotors. Figure~\ref{fig:system_overview} shows six portable and easy-to-modify modules of the simulator. 

\subsection{Quadrotor Dynamics}
We use the following quadrotor dynamics~\cite{molchanov2019sim}:
\begin{align*}
\ddot{x} &= g + \frac{\mathbf{R} f}{m}  &  \dot{\mathbf{R}} &= \bm{\omega_{\times}} \mathbf{R}  \\
\dot{\omega} &= \mathbf{I}^{-1} (\mathbf{\tau} - \omega \times (\mathbf{I} \cdot \omega))  & \tau &= \tau_{p} + \tau_{th}
\end{align*}
where $\ddot{x}$ is linear acceleration, $g$ is the gravity vector, $\mathbf{R}$ is the rotation matrix, $f$ is the total thrust force in the body frame, $m$ is the mass, $\bm{\omega_{\times}}$ is the skew matrix of the $\omega$, $\mathbf{I}$ is the inertia matrix, $\tau$ is the total torque, $\tau_{p}$ is the torque along z-axis, $\tau_{th}$ is the torque produced by motor trusts. 

The action of quadrotor $i$ is $a_{i} \in\R^4$, which represents the normalized thrust provided by each motor. Following~\cite{molchanov2019sim}, QuadSwarm models several aspects of real hardware in order to prevent policies from overfitting to the simulator and to facilitate sim2real transfer. 

\subsubsection{Motor Lag}
At timestep $t$, given actions $a^{(t)}$ from a policy sampled from an unconstrained Gaussian distribution, we constrain the actions to be in the range [0, 1] and use this to construct the normalized rotor angular velocity $\hat{u}^{(t)}$. 
\begin{align*}
\hat{f}^{(t)} &= \frac{1}{2} (\operatorname{CLIP}(a^{(t)}) + 1)  &  \hat{u}^{(t)} &= \sqrt{\hat{f}^{(t)}} 
\end{align*}
We then use a first-order low-pass filter to model motor lag 
$\hat{u}^{(t)}_{f} = \alpha_{lag} (\hat{u}^{(t)} - \hat{u}^{(t-1)}_{f}) + \hat{u}^{(t-1)}_{f}$, where $\hat{u}^{(t)}_{f}$ is the filtered rotor angular velocity, and $\alpha_{lag}$ is the motor lag time coefficient, which has been set such that the $\hat{u}^{(t)}_{f}$ satisfies 2\% settling time. 

\subsubsection{Motor Noise}
At each timestep, we sample noise from a Gaussian distribution and apply it to the motor noise value produced on the previous timestep such that $\epsilon^{(t)}_{f} = \alpha_{nd} \epsilon^{(t-1)}_{f} + \alpha_{ns} \mathcal{N}(0, 1) $, where $\epsilon^{(t)}_{f}$ is the motor noise at timestep $t$, $\alpha_{nd}$ is the decay ratio of the motor noise, $\alpha_{ns}$ is the scale factor for the motor noise, and $\mathcal{N}(0, 1)$ denotes the Gaussian distribution with zero mean and unit variance. 

% \subsubsection{Constrained Dynamics}
The final thrusts provided by each motor $f^{(t)} \in \R^4$ is constructed by the maximum thrust that each motor can provide $f_{max}$, the filtered rotor angular velocity $\hat{u}^{(t)}_{f}$, and the motor noise $\epsilon^{(t)}_{f}$. 
Specifically, $f^{(t)} = f_{max} \cdot (\hat{u}^{(t)}_{f})^{2} + \epsilon_{f}^{(t)}$.

% We constrain the maximum thrust that each motor can provide to be $f_{max}$, and use this value, the filtered rotor angular velocity $\hat{u}^{(t)}_{f}$ and motor noise $\epsilon^{(t)}_{f}$ to construct $f^{(t)} \in \R^4$, the final thrusts provided by each motor.
% \begin{align*}
%     f_{max} &= 0.25 \cdot m \cdot g \cdot \alpha_{t2w} & f^{(t)} &= f_{max} \cdot (\hat{u}^{(t)}_{f})^{2} + \epsilon_{f}^{(t)} 
% \end{align*}
% Where $\alpha_{t2w}$ is the thrust-to-weight ratio. 

% $f_{max}$ is the maximum thrust each motor can provide, $\epsilon_{f}$ is the motor noise, $\alpha_{t2w}$ is the ratio of thrust to weights.   

% $f$ is the thrust provided by each motor, $\hat u$ is the filtered normalized rotor angular velocities, which is used to simulate the motor lag, $\epsilon^{u}$ is the motor noise. 

% The transformation of policy action $a$ to thrusts is formulated as:
% \begin{align*}
%     \hat{f} &= \frac{1}{2} (a + 1)  &  \hat{u} &= \sqrt{\hat f}
% \end{align*}

% where $\hat f$ is the normalized control input. The policy action $a$ is clipped to $[-1, 1]$ before transform to $\hat f$.  

\subsection{Collision Simulation and Aerodynamics}
Modeling accurate collisions is important for learning robust collision-avoidance policies but is a non-trivial task. In this section, we outline simple collision models used by default in QuadSwarm that is implemented in a modular way and can easily be swapped with a different collision model. Although these models are simple, in~\cite{batra2022decentralized}, we demonstrated they are good enough to train successful policies.

\subsubsection{Quadrotor to Quadrotor}
When two quadrotors collide, instead of modeling complex interactions, such as whether the propellers of two quadrotors touch, we implement a simple collision model based on the linear velocity and the angular velocity.
\vspace{-2pt}
\begin{align*}
n_{col} &= \frac{x_1 - x_2}{\norm{x_1 - x_2}_2} & \Tilde{v} &= (v_{2} \cdot n_{col} - v_{1} \cdot n_{col}) \cdot n_{col} \\
v_{1} &\leftarrow \alpha_{1} (v_{1} + \Tilde{v} + \epsilon_{v1})  &  v_{2} &\leftarrow \alpha_{2} (v_{2} - \Tilde{v} + \epsilon_{v2})  \\
\omega_{1} &\leftarrow \omega_{1} + \epsilon_{\omega 1}  & \omega_{2} &\leftarrow \omega_{2} + \epsilon_{\omega 2}
\end{align*}
Where $x_1, x_2$ are the positions of two quadrotors, $v_1, v_2$ are the linear velocity of two quadrotors, $\alpha_1, \alpha_2$ are the linear velocity decay factor of two quadrotors, $\epsilon_{v1}, \epsilon_{v2}$ are the linear velocity noise of two quadrotors, and $\epsilon_{\omega 1}, \epsilon_{\omega 2}$ are the angular velocity noise of two quadrotors.
% \james{I was super confused reading this at first because ``collision model'' meant something geometric that takes two quadrotor states and returns true iff they are in collision. But now I see this is actually about what happens to the states after they collide. Please state this explicitly. Also, you should either use notation like tilde, bar, etc. for the post-collision states, or use $\gets$ instead of $=$.}

\subsubsection{Quadrotor to Wall or Ceiling}
The collision model between a quadrotor and walls or ceiling is the same as the quadrotor-quadrotor collision model, except that the collision updates are only applied to the quadrotor.

\subsubsection{Quadrotor to Ground}
We consider two situations of quadrotor interaction with the ground. When the quadrotor hits the ground we set the linear velocity, angular velocity, and acceleration to zero, regenerate the rotation matrix by setting the normal vector of the quadrotor upward, and reset all momenta. When the quadrotor is on the floor, and the thrust is not enough to allow the quadrotor to take off, we arrest motion on the floor with sufficiently high friction. When the linear velocity of the quadrotor is $0$, the friction direction is opposite to the thrust force direction in the $xy$ plane, and the final force function is: $ f_{xy} \leftarrow \max(f_{xy} - \mu (mg - f_z), 0)$. 
When the linear velocity is bigger than $0$, the friction direction is opposite to the velocity direction in the $xy$ plane, and the final force function is: $f_{xy} \leftarrow f_{xy} - \mu (mg - f_z)$. In functions above, $f_{xy}$ is the thrust force in the $xy$ plane, $f_z$ is the thrust force in $z$ axis, $\mu$ is the friction coefficient, and $g$ is the gravity constant.

% The first situation is the linear velocity of the quadrotor is $0$, the friction direction is opposite to the thrust force direction in the $xy$ plane:
% \begin{align*}
% \bar f_{xy} &= \max(f_{xy} - \mu (mg - f_z), 0) & acc_{xy} &= \frac{\bar f_{xy}}{m} & acc_z = 0 
% \end{align*}

% The second situation is the linear velocity of the quadrotor is bigger than $0$, then, the friction direction is opposite to the velocity direction in the $xy$ plane:
% \begin{align*}
% \bar f_{xy} &= f_{xy} - \mu (mg - f_z) & acc_{xy} &= \frac{\bar f_{xy}}{m} & acc_z = 0 
% \end{align*}

% \begin{align*}
% \bar f_{xy} &= \max(f_{xy} - \mu (mg - f_z), 0) & acc_{xy} &= \frac{\bar f_{xy}}{m} & acc_z = 0 
% \end{align*}

% where $\bar f_{xy}$ is friction force in the $xy$ plane, $f_{xy}$ is the total thrust force in the $xy$ plane, $\mu$ is the friction coefficient, $mg$ is the gravitational force of the quadrotor, $f_z$ is the thrust force in the $z$ axis, $acc_{xy}, acc_z$ is the acceleration in the $xy$ plance and in the $z$ axis. 

% \james{Details on friction model?}

\subsubsection{Downwash}
%When there are one or multiple quadrotors flying above the quadrotor, the airflow above the quadrotor will be changed, which can use a disturbance to the quadrotor. 
Our downwash model is a simplified version of the model proposed in~\cite{shi2020neural}. We only model downwash effects when two quadrotors overlap in the $xy$ plane and within a pre-defined distance along the $z$ axis.
\begin{align*}
\ddot{x} &= k_1 (k_2 \delta_{pos} + b_1) + \epsilon_d & \dot{\omega} &= \epsilon_{\omega d}
\end{align*}
Where  $\delta_{pos}$ is the relative distance between quadrotors, $\dot{\omega}$ is the change rate of angular velocity, which is used to simulate the aerodynamic disturbances, and $k1, k2, b1$ are constants, $\epsilon_d, \epsilon_{\omega d}$ are Gaussian noise.  

\subsection{Observations}
The observations of quadrotor $i$ are: 
\begin{align*}
[\delta_{xi}, v_i, R_i, \omega_i, [\Tilde{x_{i1}}, \Tilde{v_{i1}}, ..., \Tilde{x_{iK}}, \Tilde{v_{iK}}]]
\end{align*}
where $\delta_{xi}$ represents the relative position between the quadrotor $i$ and its goal, $\Tilde{x_{i1}}, \Tilde{v_{i1}}$ represent the relative position and relative velocity to the closest quadrotor, $\Tilde{x_{iK}}, \Tilde{v_{iK}}$ represent the relative position and relative velocity to the Kth closest quadrotor. K is a hyperparameter. In the single quadrotor environment, $K$ is set to 0.

To increase zero-shot sim-to-real transfer ability, we add sensor noise to the observations~\cite{molchanov2019sim}: 
\begin{align*}
\epsilon_{x} = \it{U}(0, 5e^{-3}) && \epsilon_{v} = \it{U} (0, 1e^{-2}) && \epsilon_{\omega} = \mathcal{N}(0, 1.75e^{-4})
\end{align*}
where $\it {U}$ represents the uniform distribution, $\mathcal{N}$ represents the Gaussian distribution, $\epsilon_{x}$ is the position noise, $\epsilon_{v}$ is the linear velocity noise, $\epsilon_{\omega}$ is the angular velocity noise.

% The action of quadrotor $i$ is $a\in\R^4$, which represents the normalized thrust provided by each motor. 
\subsection{Training Scenarios}
To design diverse training scenarios, we use the quadrotor team's goals to construct several geometric formations, including a circle, grid, sphere, cylinder, and cube. We use this pool of geometric formations to design three groups of training scenarios.

\subsubsection{Static formations} 
Uniformly sample a geometric formation from the pool and randomly place it in the room. 

\subsubsection{Dynamic formations}
Change the positions and/or the geometric formation of goals after a random period of time within an episode. There are four variants: 
\begin{itemize}
  \item Dynamic goals: regenerate the positions and the geometric formation of goals after a random period of time.  
  \item Swap goals: keep the geometric formation but shuffle the positions of goals after a random period of time.
  \item Shrink $\&$ Expand: keep the geometric formation of goals, but change the formation size over time.
  \item Swarm-vs-Swarm: split quadrotors into two groups, and fix the formation center of each group. After a random period of time, resample the formation shape and swap the goals of the two groups.
\end{itemize}

\subsubsection{Evader Pursuit}
Quadrotor(s) pursue one moving goal. We parameterize the trajectories in two ways - using a 3D Lissajous curve, and randomly sampled consecutive points connected by Bezier splines, respectively. 

\subsection{Reward Components}
We provide diverse reward components in the simulator. There are two groups of reward components. One is based on the quadrotor's state, and the other is based on the interactions with other objects. All $\alpha$ below are constants.

\noindent Quadrotor State:
\begin{align*}
r^{(t)}_{pos} &= \alpha_{pos} \norm{\delta^{(t)}_{xi}}_2 & r^{(t)}_{vel} &= \alpha_{vel} \norm{v^{(t)}}_2  \\
r^{(t)}_{ori} &= \alpha_{ori} R^{(t)}_{22}  &  r^{(t)}_{spin} &= \alpha_{spin} \norm{\omega^{(t)}}_2\\
r^{(t)}_{act} &= \alpha_{act} \norm{f^{(t)}}_2  &  r^{(t)}_{\delta act} &= \alpha_{\delta act} \norm{f^{(t)} - f^{(t-1)}}_2  \\
r^{(t)}_{rot} &= \alpha_{rot} \frac{tr(R^{(t)}) - 1}{2}  &  r^{(t)}_{yaw} &= \alpha_{yaw} R^{(t)}_{00}
\end{align*}
where reward components based on the distance to the goal, linear velocity, the normal vector in the z-axis, angular velocity, actions, change of actions, rotation, and yaw.
% \james{are these norms definitely not squared? I can't remember.}

\noindent Interaction with Other Objects: We use a weighted combination of indicator functions for the conditions when the quadrotor hits the floor, stays on the floor, hits a wall, hits the ceiling, or hits other quadrotors. We also use a weighted combination of the relative distance between quadrotors for the condition when quadrotors are close to each other.  

% We use the indicator function to provide instance penalties upon collision with other quadrotors or room. We also provide 
% \begin{align*}
% r_{on\_floor} &= \alpha_{on\_floor} \1_{on\_floor} & r_{hit\_floor} &= \alpha_{hit\_floor} \1_{hit\_floor} \\
% r_{hit\_wall} &= \alpha_{hit\_wall} \1_{hit\_wall} & r_{hit\_ceil} &= \alpha_{hit\_ceil} \1_{hit\_ceil}
% \end{align*}
% \begin{align*}
% r_{hit\_quad} &= \alpha_{hit\_quad} \1_{hit\_quad} \\
% r_{near\_quad} &= \alpha_{near\_quad} \sum_{j=1}^{K} max(1 - \frac{\norm{{\Tilde{x_{ij}}}}}{\alpha_{prox}}, 0)
% \end{align*}

% where we provide reward components given if the quadrotor is on the floor, hit the floor, hit the wall, hit the ceiling, hit with other quadrotors, and close to other quadrotors. The indicator function $\1$ is equal to 1 when the interaction happens at timestep $t$. $\Tilde{x_{ij}}$ represents the relative distance between quadrotor i and its the jth closest quadrotor.
% \james{Defining math notation for the indicator-based reward terms doesn't really add anything here and takes up space. The verbal description is enough -- you can just say ``a weighted combination of indicator functions for the conditions \dots (hit floor, etc.)''}

\subsection{Reinforcement Learning Library Interface}
We integrate Sample Factory~\cite{petrenko2020sample}, a fast RL library, with QuadSwarm to decrease the wall-clock training time.
% \james{Acronym definition should always be at the site of first use, not way down here.}
Sample Factory supports synchronous and asynchronous modes of policy proximal optimization (PPO) algorithms. For multi-agent RL, it currently supports Independent PPO. 
% \james{use proper \LaTeX\ quotes.}

\section{Simulation Speed}
To balance speed, readability, and flexibility, we decide to: $(i)$ use Python to implement the minimum requirements of physics simulation and rendering, $(ii)$ use Numba~\cite{lam2015numba}, a just-in-time compiler that is able to translate Python and NumPy code into machine code to speed up physics simulations, and $(iii)$ decouple rendering from physics simulations.

\begin{figure}[htbp!]
\centering
    \vspace{-6pt}
    \includegraphics[width=0.36\textwidth]{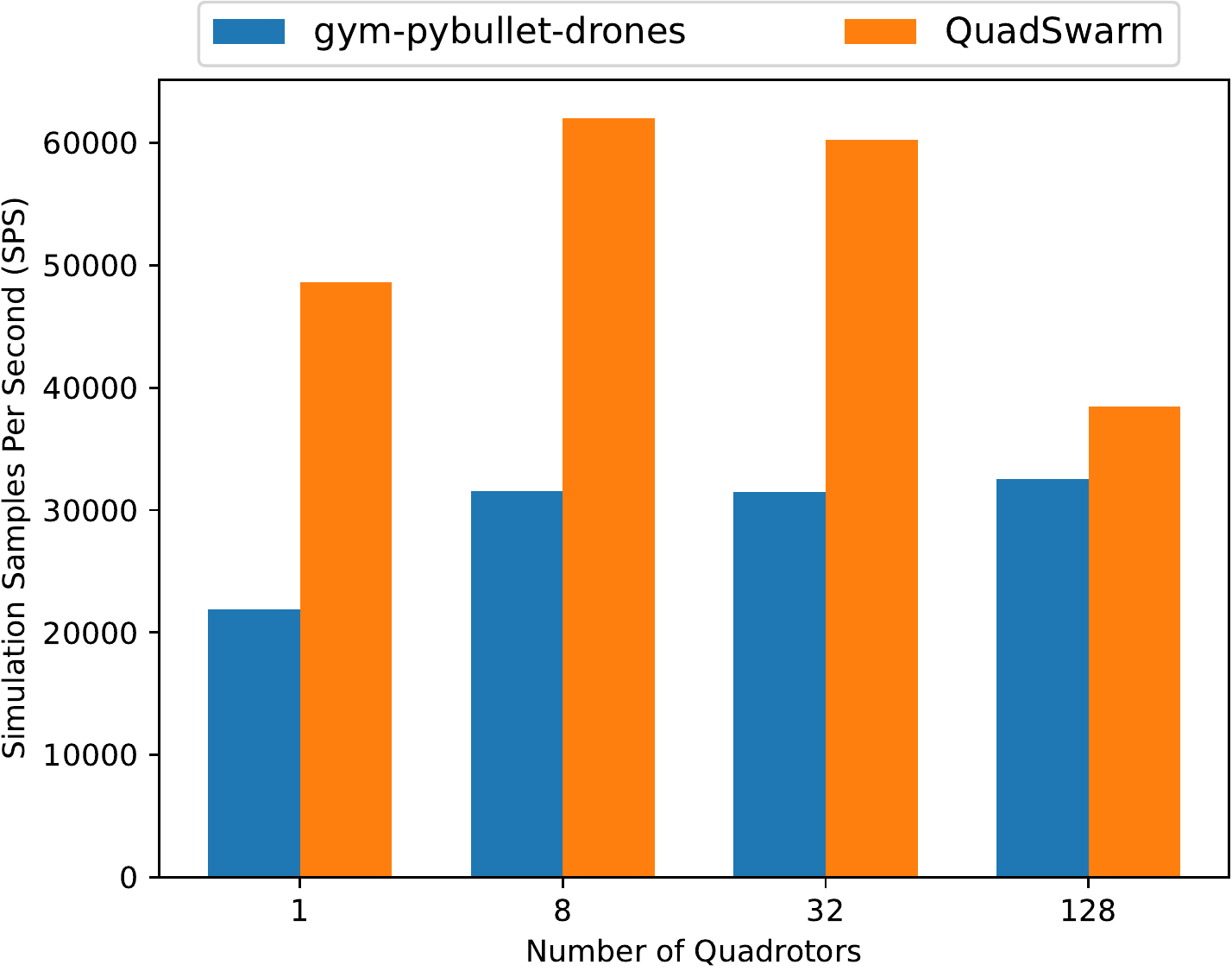}
    \vspace{-10pt}
    \caption{Simulation Speed: gym-pybullet-drones vs QuadSwarm}
    \label{fig:sim_speed}
    \vspace{-10pt}
\end{figure}

We evaluate simulation speed on a machine with AMD Ryzen 7 2700X CPU (16 CPU cores). To fairly compare QuadSwarm with gym-pybullet-drones, we set both simulators with 100 Hz control frequency, 200 Hz simulation frequency, and 15 seconds episode duration time. In an environment with multiple quadrotors, each quadrotor has the same observation space, thus QuadSwarm receives multiple samples per simulation step. 
% each quadrotor has the same observation space. 
%This makes the simulator gets multiple samples per simulation step and might speed up simulation even with collisions.    

Fig.~\ref{fig:sim_speed} shows the simulation speed comparison between gym-pybullet-drones and QuadSwarm. In a single quadrotor setting, QuadSwarm approaches 48,589 SPS - $\sim$2.2x faster than gym-pybullet-drones. 
 With multiple quadrotors and collision simulation, QuadSwarm approaches the fastest simulation speed, 62,042 SPS, when the number of quadrotors is eight - $\sim$2.0x faster than gym-pybullet-drones.

% On the other hand, the simulation speed of QuadSwarm decreases as the number of quadrotors increases, while gym-pybullet-drones does not.

%when compare the simulation speed between the environment with 32 quadrotors and 128 quadrotors. 

% This is due to the larger computation time of obtaining neighbor observations in QuadSwarm, which uses relative position and relative velocity, whereas gym-pybullet-drones uses Boolean adjacency distance matrix.

% The control frequency is 100 Hz, the simulation frequency is 200 Hz, and the episode duration time is 15 seconds. the simulation speed approaches 37608 FPS on a single quadrotor, approaches 50066 FPS on 8 quadrotors, and 47702 FPS on 128 quadrotors.  
% \james{See earlier comment about FPS meaning.}

\section{Examples}
We used QuadSwarm as the main simulation platform in two projects that demonstrated the transfer of learned control policies on single and multiple quadrotors. For a single quadrotor~\cite{molchanov2019sim}, we show how to learn a policy to stabilize multiple different quadrotors with domain randomization. For multiple quadrotors~\cite{batra2022decentralized}, we show how to learn a policy to control up to 128 quadrotors to approach their goals while avoiding collisions in diverse scenarios.

\section{Conclusions}
We describe QuadSwarm, a simulator for Deep RL research on single and multi-quadrotor control policies and their sim2real transfer to real hardware. We demonstrate how QuadSwarm integrates five key ingredients:
$(i)$ a reasonable physics model of Crazyflie 2.x, with domain randomization to account for unmodeled effects;
$(ii)$ per-rotor thrust control; 
$(iii)$ fast, high parallelization, and scaling with additional compute; 
$(iv)$ a diverse collection of learning scenarios for single and multi-quadrotor teams;
$(v)$ 100$\%$ written in Python. 
% $(i)$ an accurate physics model of an existing platform and support for sufficient domain randomization for sim2real transfer $(ii)$ fast single-threaded throughput and high parallelization capability, 
% and $(iii)$ a diverse collection of training scenarios. 
Our experiments suggest that QuadSwarm can be used to create robust quadrotor policies that successfully deploy to real hardware and that it is a useful and promising tool that will accelerate research in robust single and multi-quadrotor control policies for agile flight. We are working on extending QuadSwarm to support multiple obstacles, providing more accurate aerodynamic effects, and integrating with additional Deep RL libraries, such as PyMARL2~\cite{hu2021rethinking}.

\bibliographystyle{IEEEtran}
\bibliography{IEEEexample}
\end{document}